\documentclass[a4paper,11pt]{article}
\usepackage[utf8]{inputenc}
\usepackage[british]{babel}
\usepackage[mono=false]{libertine}
\usepackage[T1]{fontenc}
\usepackage[top=2.5cm, bottom=2.5cm, left=2cm, right=2cm]{geometry}
\usepackage[font=small]{caption}
\usepackage{subcaption}
\usepackage{cite}
\usepackage{enumitem}
\usepackage{titlesec}
\usepackage[usenames, dvipsnames]{color}
\usepackage[titletoc]{appendix}
\usepackage[pdftex]{graphicx}
\usepackage{array}
\usepackage{url}
\usepackage{mathtools}
\usepackage{amssymb,amsfonts,amsmath,amsthm}
\usepackage{dsfont}
\usepackage{bm}
\usepackage[perpage,hang,flushmargin]{footmisc} %Reset the footnote counter each pag
\usepackage{enumitem}
\usepackage{booktabs}       % professional-quality tables
\usepackage{nicefrac}       % compact symbols for 1/2, etc.
\usepackage{microtype}      % frac{
\usepackage{tikz}
\usepackage[ruled,vlined]{algorithm2e}
\usepackage{wrapfig}
\usepackage{xr,xr-hyper}
\usepackage{cases}
\usepackage{setspace}

\usepackage[pdfpagemode=UseNone,bookmarksopen=false,colorlinks=true,urlcolor=blue,citecolor=Red3,citebordercolor=blue,linkcolor=blue]{hyperref}
\usepackage[capitalize]{cleveref}

\usepackage{natbib}

\usepackage{graphicx,amsmath,amssymb,dsfont,bm,float,xcolor,ipa}
\definecolor{darkred}{RGB}{180,0,0}

\usepackage{hyperref}
\hypersetup{
  colorlinks=true,breaklinks,
  linktoc=section,
  linkcolor=darkred,
  linkbordercolor=white,
  citecolor=darkred,
  citebordercolor=darkred,
  urlcolor=darkred,
  pdfborder = {0 0 1}
}

\title{A Simple Model of Inference Scaling Laws}
\author{}

\author{%
  Noam Levi\\
  École polytechnique fédérale de Lausanne (EPFL)\\
  CH-1015 Lausanne \\
  \texttt{noam.levi@epfl.ch} \\
}

\begin{document}
\date{}

\maketitle

\begin{abstract}
Neural scaling laws have garnered significant interest due to their ability to predict model performance as a function of increasing parameters, data, and compute. In this work, we propose a simple statistical ansatz based on memorization to study scaling laws in the context of inference, specifically how performance improves with multiple inference attempts. We explore the coverage, or pass@k metric, which measures the chance of success over repeated attempts and provide a motivation for the observed functional form of the inference scaling behavior of the coverage in large language models (LLMs) on reasoning tasks. 
We then define an "inference loss", which exhibits a power law decay as the number of trials increases, and connect this result with prompting costs. We further test our construction by conducting experiments on a simple generative model, and find that our predictions are in agreement with the empirical coverage curves in a controlled setting. Our simple framework sets the ground for incorporating inference scaling with other known scaling laws.

\end{abstract}

\section{Introduction}

Advancements in deep learning have demonstrated that the performance of neural networks scales predictably as a function of model size, data size, and computational resources~\citep{hestness2017deep, kaplan2020scaling, rosenfeld2020a, henighan2020scaling}. These trends, known as \textit{neural scaling laws}, have motivated research into understanding how scaling influences model performance in a range of domains, in particular, Large Language Models (LLMs)~\citep{brown2020language, hoffmann2022empirical}.

However, scaling during \textit{inference}---the process by which a trained model makes predictions on new data---has received less attention. Recent works have shown empirically that LLMs can gain substantial benefits from repeated prompts to perform better on difficult tasks such as coding and formal proofs, where verification of the correct answer can be done~\citep{brown2024largelanguagemonkeysscaling, snell2024scalingllmtesttimecompute, Bansal2024SmallerWY}. These works demonstrate that the performance of weaker models can be amplified without further training, by simply repeating inference trials. A natural question then arises:
\begin{align}
\nonumber
    \text{\it Can we interpret, or predict the inference scaling behavior of a model with repeated attempts?}
\end{align}

To answer this question, we propose a simple toy model that isolates the \textit{inference scaling laws} which dictate how certain performance metrics improve as a function of the number of inference attempts. Inspired by the work of \citet{hutter2021learningcurvetheory}, which introduced a model to study scaling behavior for memorization and generalization, we devise a simple setting to capture the effect of repeated inference attempts, focusing on the \textit{coverage} metric, also known as \textit{pass@k}.

In this work, we present analytical predictions for coverage from a probabilistic perspective and demonstrate how inference improves with the number of repeated trials in a predictable way, which matches the observed behavior in~\cite{brown2024largelanguagemonkeysscaling} and~\cite{snell2024scalingllmtesttimecompute}.  
We use two different approaches to obtain the predicted pass@k, and highlight the connection between coverage and total inference cost. Additionally, we define a simple "inference loss", similar to the familiar test loss, but allowing for repeated trials, and predict its scaling.
Lastly, we verify our predictions empirically by appealing to a simple generative model, a Variational Autoencoder (VAE), which is trained to generate reconstructions of its training data by sampling from a latent space with an associated temperature, mimicking some of the complex generative properties of LLMs.
Given that our results are isolated from the effects of other neural scaling laws, they could be incorporated into a broader exploration to find the optimal train/inference point. In particular, we hope this work sets the ground for exploring the optimal trade-off between training and inference attempts, such that the total performance, as well as cost, is minimized.

\section{Related Work}
Scaling laws for neural networks have been extensively studied in recent years. Empirical research has shown that error rates decrease predictably as a function of increased data, parameters, and compute, following power-law relationships. Notable contributions in this space include work by~\citet{kaplan2020scalinglawsneurallanguage}, who demonstrated consistent scaling behavior across language models, and~\citet{henighan2020scalinglawsautoregressivegenerative}, who extended these ideas to multimodal models, as well as~\citet{cabannes2024scalinglawsassociativememories, maloney2022solvable, bordelon2020spectrum, spigler2020asymptotic,caponnetto2007, steinwart2009, fischer2020,cui2021generalization,levi2023universalstatisticalstructurescaling, levi2024underlyingscalinglawsuniversal,nam2024exactlysolvablemodelemergence} who studied scaling laws for solvable yet sufficiently complex models, ranging from generalized linear regression on random feature models to kernel ridge regression, connecting them to results from random matrix theory and the underlying properties of the data.  
While most scaling laws focus on training, the study of \textit{inference scaling} remains under-explored. Our work tries to fill this gap by studying how performance improves with repeated inference attempts.

\section{Memorizing Ansatz}

In the following, we first briefly review the simplest model which produces the known data scaling law prediction by appealing to a memorizing construction, then consider our proposed model for repeated inference attempts.

% \subsection{Scaling with Samples in the Hutter Model}

The \textit{Hutter model}~\citep{hutter2021learningcurvetheory} is a probabilistic framework originally introduced to study the scaling behavior of learning curves, focusing on the relationship between memorization and generalization. It assumes a model which perfectly memorizes features during training, allowing it to correctly match sets of features and labels $\{i, y_i\}$, such that only unseen features can incur an error.
The set of features is assumed to follow a Zipf power law decay with parameter $\alpha$, where the probability of encountering feature $i$ is $\theta_i \propto i^{-1-\alpha}$ and decreases with its rank. For $n$ training samples, the expected single feature error is
% given by
\begin{align}
\label{eq:hutter}
\mathcal{E}_n
=
\mathbb{E}
[E_i] = 
\sum_{i=1}^\infty 
\theta_i (1 - \theta_i)^n
\approx n^{-\beta}, \qquad \beta = \frac{1}{1+\alpha}.
\end{align}
\cref{eq:hutter} captures the average likelihood of encountering and labeling the feature incorrectly after $n$ training samples. 
Similar to this model, we will adopt the idea of memorization as a surrogate for training, but will depart from the finite training set assumption as a basis for test errors. 
Since we are only interested in successes at the single sample level, we take the index $i$ to represent the samples themselves
% , rather than features 
from here on.

\subsection{Perfect Sample Memorization, Imperfect Predictions}

In contrast to the Hutter model, we focus on a scenario where \textit{all samples have been memorized}, hence there is no notion of test error coming from unseen data. Instead, failure during inference or data generation could arise from the need to follow a sequence of steps to reach the correct answer. 

Concretely, we consider a joint model of memory $M$ and inference $I$. 
The memory is $M\in \mathcal{M}:=\mathcal{X} \to \mathcal{Y}$,
where the samples are identical to the labels $\{x_i\}_{i=1}^n=\{y_i\}_{i=1}^n$, and corresponds to a model which simply learns to memorize its training data. 
The inference model $I\in \mathcal{H}:=\mathbb{N} \to M$ takes an input index $i$, which serves as a proxy for a particular prompt or task, and should produce the associated memorized label (sample) $y_i$ by recalling it from the memory $M$. The inference model is taken to be imperfect in its retrieval, and subject to some error $\epsilon$, it makes predictions
\begin{align}
    I(i)= 
    \left\{
    \begin{array}{cc}
        y_i+\epsilon, & \text{with probability }  p_i,\\
        y_i, & \text{with probability }   1-p_i .\\
        \end{array}\right.
\end{align}
% These errors may accumulate due to the internal probability distribution learned by the model and the sampling temperature, leading to failures.
At this point, we only accept perfect reconstructions, such that the performance of the model on a single sample is measured simply by
\begin{align}
    A(i) = \mathbf{1}_{\{y_i\}}(I(i)),
\end{align}
where $\mathbf{1}_{\mathcal{Y}}(x)=1~\text{if }x\in \mathcal{Y}~\text{else } 0$ is the indicator function.

We start from the general case where each sample has an unknown failure probability $p_i\in[0,1]$ at inference, and we are interested in the probability of at least one successful generation of a sample over $k$ attempts. We further assume that if a sample has been correctly generated during inference, we have access to a perfect verification method, which can determine if a single sample is correct among $k$ possible candidates.

Assuming the attempts are independent and identically distributed (i.i.d.), the probability of failing on all $k$ trials for a sample with failure probability $p_i$ is simply $\mathds{P}(k~\text{failures on sample}~i)=\prod_{t=1}^{k}  (1-A_t(i)) =  p_i^k$, where $t$ is the trial index. Therefore, the probability of at least one success in $k$ trials averaged over the entire dataset of size $n$, known as \textbf{pass@k}, is given by
\begin{align}
\text{pass@k} = 1 - \frac{1}{n}\sum_{i=1}^{n}\prod_{t=1}^{k}(1-  A_t(i) )
=
1 - \frac{1}{n}\sum_{i=1}^{n}
p_i^k. 
\end{align}
This will be the metric we use to describe inference accuracy for the rest of this work.

\subsection{Power Law Distributions Predict Pass@k for LLMs}

\begin{figure}[t!]
    \centering
    \includegraphics[width=.98\linewidth]{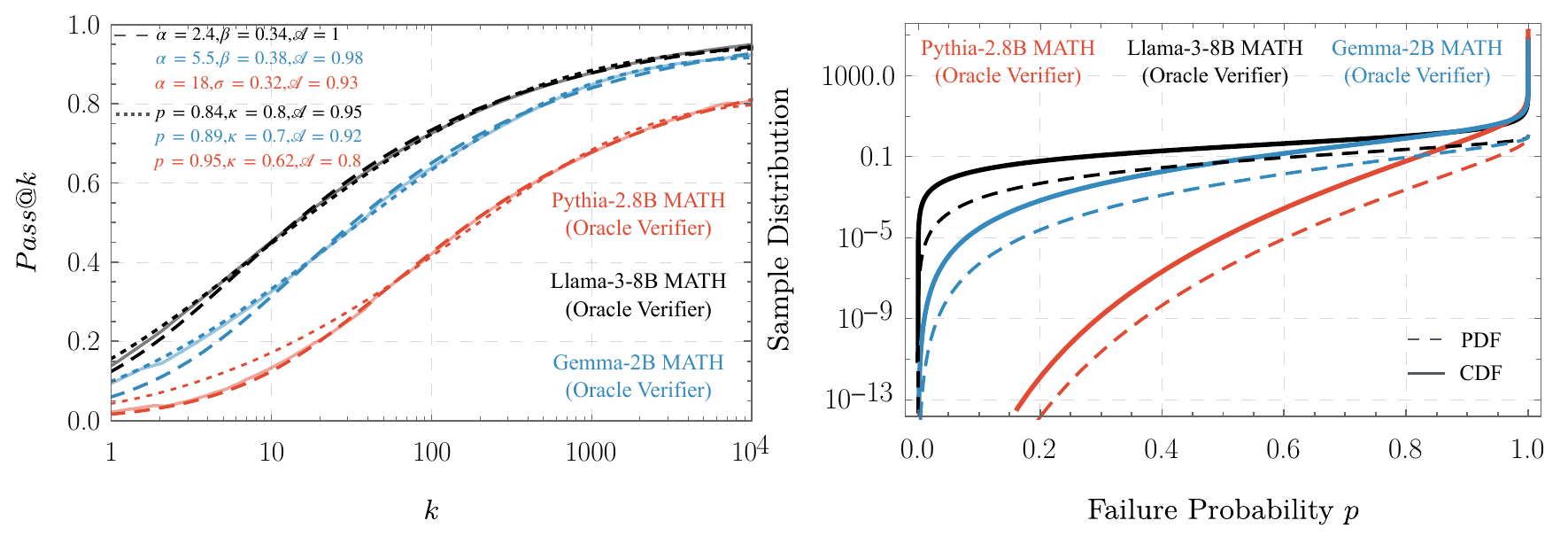} 
    \caption{{\bf{Pass@k and failure distribution curves for various LLMs on difficult tasks, against theoretical scaling predictions}}. 
    {\it Left:} 
    The relationship between pass@k and the number of samples for several coding and maths tasks for different models, as described in~\cite{brown2024largelanguagemonkeysscaling}, compared with the analytical predictions presented in~\cref{eq:main_result,eq:pass@k}. The {\it solid} curves are data, while the {\it dashed} curves are the predictions from \cref{eq:main_result}, where $\alpha$ and $\beta$ correspond to the concentration of easy and hard problems, respectively. The {\it dotted} curves are the results of \cref{eq:pass@k}. The functional form in both cases captures well the LLM pass@k curves for various models, by adjusting $\alpha,\beta$ or $p,\kappa$.
{\it Right:} The $\text{Beta}(\alpha,\beta)$ distributions for the failure probabilities are shown for the different models. We can see that most of the questions are "difficult", while the existence of a left tail implies that more trials are required to obtain a correct answer. 
}
    \label{fig:appx_results}
\end{figure}

To construct the distribution of failure probabilities $p_i$, we assume that different samples may have different inference complexity levels, incorporating some "easy" and some "difficult" samples with respect to the inference model. 
One way to model the different complexities is done by appealing to the so called Beta distribution. We think of $p=p_i$ as a random variable, drawn from $p\sim\text{Beta}(\alpha,\beta)$, where $\alpha\in\mathbb{R}^+$ controls how often we encounter "easier" problems, where smaller $\alpha$ pushes the distribution mass towards zero, while $\beta\in\mathbb{R}^+$ dictates how often we encounter "harder" problems. Namely, a lower $\beta$ parameter increases the distribution mass towards the right tail (high failure probabilities). The $\text{Beta}(\alpha,\beta)$ PDF is given explicitly by
\begin{align}
    \text{Beta}(\alpha,\beta; p) =
    \frac{p^{-1+\alpha } (1-p)^{-1+\beta}}{B(\alpha ,\beta )},
\end{align}
where $B(\alpha ,\beta )$ is the Euler beta function. 
We can therefore compute the pass@k metric by simply averaging over the failure distributions as
\begin{align}
\label{eq:main_result}
\text{pass@k} 
&=
\mathcal{A}\times\left(1 - \frac{1}{n}\sum_{i=1}^{n} p_i^k\right)
\approx
\mathcal{A}\times\left(1-\langle p^k \rangle\right)
=
\mathcal{A}\times\left(1-\int_{0}^1 dp p^k \frac{p^{-1+\alpha} (1-p)^{-1+\beta }}{B(\alpha ,\beta )}\right)
\\ \nonumber
&=
\mathcal{A}\times\left(
1-\frac{\Gamma (\beta ) \Gamma (k+\alpha )}{B(\alpha ,\beta ) \Gamma (k+\alpha +\beta )} 
\right)
\underset{k\to \infty}{\approx}
\mathcal{A}\times\left(1-
\frac{\Gamma (\beta ) k^{-\beta }}{B(\alpha ,\beta )}
\right),
% 1-\frac{1}{\sqrt{2 \pi } \sigma }
% \int_0^\infty
% e^{-
% % \left(
% k e^{-L}
% % e^{
% -\frac{(L -\mu )^2}{2 \sigma ^2}
% % \right)
% }
% % }
% d L
% \\ \nonumber
% &\approx
% 1-\frac{\exp \left(-\frac{\mathcal{W}\left(k e^{-\mu } \sigma ^2\right)^2}{2 \sigma ^2}-k e^{-\mathcal{W}\left(k e^{-\mu } \sigma ^2\right)-\mu }\right)}{\sigma  \sqrt{k e^{-\mathcal{W}\left(k e^{-\mu } \sigma ^2\right)-\mu }+\frac{1}{\sigma ^2}}}
% , \,
\end{align}
where $\Gamma(z)$ is the Euler gamma function, and $\langle . \rangle$ indicates averaging over the $p$ distribution. Since real world models such as LLMs are not perfect memorizers by our definitions, we also add an overall factor describing their maximal pass@k as $\mathcal{A}$.
% This is yet another example where power-law scaling emerges as a universal description of the underlying phenomenon.
In~\cref{fig:appx_results}, we show that the functional form of $\text{pass@k}$ given in~\cref{eq:main_result} is a good approximation for the empirical pass@k curves for various LLMs on mathematical and coding tasks, as reported in~\citet{brown2024largelanguagemonkeysscaling}.

\begin{figure}[t!]
    \centering
    \includegraphics[width=.9\linewidth]{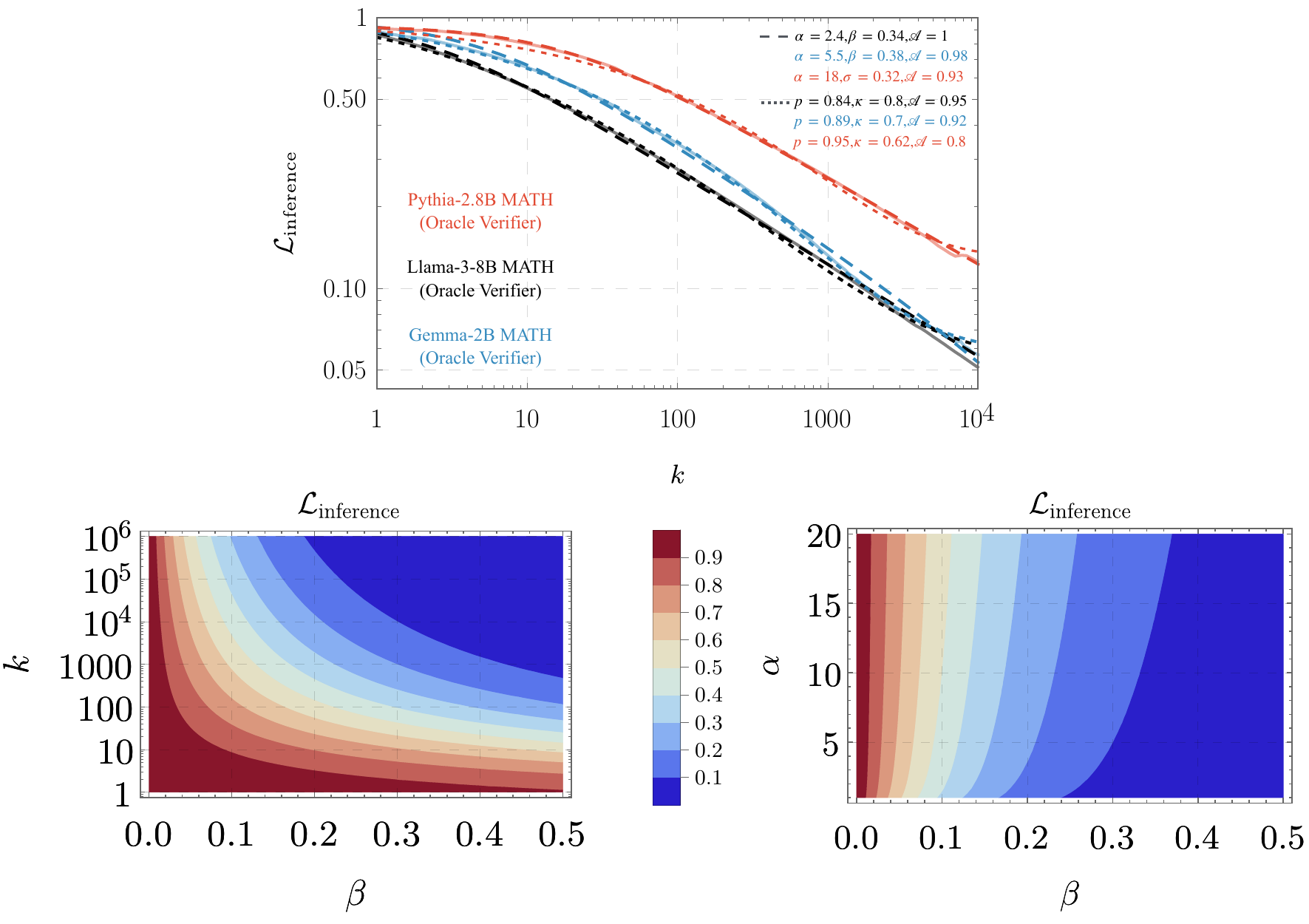} 
    \caption{{\bf{Inference attempts loss $\mathcal{L}_{\text{inference}}(k)$ for repeated attempts on the memorizing model}}.
    {\it Top:} Inference loss as a function of trials for the LLM experiments in~\cite{brown2024largelanguagemonkeysscaling}.
    {\it Bottom:} Inference loss for different $\beta$ and $k$ values. Different colors indicate inference loss values at fixed $\alpha=5$ ({\it left)} and at fixed  $k=10^4$ ({\it right)}, illustrating the behavior of \cref{eq:loss}.
}
    \label{fig:loss}
\end{figure}

We see that at large $k$, we find a power law decay of the effective average failure probability, a common feature in neural scaling laws. If we then define the {\bf inference loss} $\mathcal{L}_{\text{inference}}(k)$ as the expectation with respect to the sample distribution over errors, our results for pass@$k$ correspond to

\begin{align}
\label{eq:loss}
    \mathcal{L}_{\text{inference}}(k)
    \equiv
    \mathbb{E}(\text{Error in $k$ trials})
    =
   \mathbb{E} (\mathcal{A}\times p^k)
    \approx
    \mathcal{A}\times
    \frac{\Gamma (\beta ) \Gamma (k+\alpha )}{B(\alpha ,\beta ) \Gamma (k+\alpha +\beta )} 
    \underset{k\to \infty}{\approx}
    \mathcal{A}\times\frac{\Gamma (\beta ) k^{-\beta }}{B(\alpha ,\beta )}.
\end{align}
This result implies that the model test loss with repeated inference steps will decrease mainly depending on the value of the exponent $\beta$. Intuitively, it means that for a fixed $\alpha$ parameter, the harder the questions appear to the model, the more inference attempts are required to reach a low loss. We illustrate this behavior in~\cref{fig:loss}.

\subsubsection{Interpretability}

The result in~\cref{eq:loss} not only allows one to predict a threshold for inference in order to get an increase in model performance on difficult problems, but could also offer some interpretation for the difficulty of tasks with respect to the trained model. 
We can gain some insights regarding the real-world models by fitting the pass@k metric according to~\cref{eq:main_result} to the ones given in~\cite{brown2024largelanguagemonkeysscaling}, and attempt to interpret the properties of the test data from the parameters $\alpha$ and $\beta$.
In particular, note that the ratio $\frac{\alpha}{\alpha+\beta}$ gives the mean of the Beta distribution, which represents the average failure probability across the samples.
If $\alpha>\beta$, the mean failure probability is high (i.e., most samples are harder). On the other hand, if $\alpha<\beta$, the mean failure probability is low, implying that most samples are easy. 
Furthermore, the denominator $\alpha+\beta$ governs the concentration of the distribution, such that when 
$\alpha+\beta$ is large, the failure probabilities $p_i$ are more tightly clustered around the mean (more homogeneity in difficulty), and conversely for small $\alpha+\beta$.

From \cref{fig:appx_results}, we can see that the typical values for the "harder" problem parameter $\beta\sim0.35$, while the tail parameter is $2<\alpha<20$. 
This implies that most of the problems in the datasets used to measure the pass@k curves were indeed difficult in terms of the model's perception, while the existence of a left tail implies that the easy samples are covered quickly with less than $100$ trials. This is reflected in the right panel of~\cref{fig:appx_results}.

Another point of interest is that the average performed in~\cref{eq:main_result} can be thought of as a Laplace transform from the variable $\sigma = \log{(1/p)}$ to the trials space $k$, in the sense that
\begin{align}
\tilde{f}(k)
    =
    \langle p^k\rangle
    =
    \int_0^\infty
    d\sigma
    e^{-\sigma k }
    \frac{e^{-\alpha  \sigma } \left(1-e^{-\sigma }\right)^{-1+\beta }}{B(\alpha ,\beta )}
        =
    \int_0^\infty
    d\sigma
    e^{-\sigma k }
    f(\sigma).
\end{align}
This interpretation implies that it is possible to derive the probability distribution function of the samples in terms of their perceived difficulty by performing the inverse transform. In particular, given an empirical pass@k metric obtained for a given model, the inverse transform on $\tilde{f}(k) = (\mathcal{A}-\text{pass@}k)/\mathcal{A}$ will yield the perceived difficulty PDF. Potentially, such a procedure can be used to identify "difficult" and "easy" questions and construct improved fine-tuning algorithms by choosing training samples biased towards the "difficult" but obtainable tasks.

% Interestingly, the curves are well approximated by taking the simplest case of constant probabilities $p_i=p$ for all samples. 

\subsection{Correlated Trials and Effective \( k \) Approach}

In the previous section, we showed that correlated samples drawn iid from a varying complexity distribution can effectively describe the pass@k metric for memorizing models. Here, we take a converse approach, where do not assume that samples are correlated through their failure rate distribution, but instead that {\it trials} themselves are correlated. 

One can conjecture that dependencies between trials arise due to the internal model structure and the data itself. To capture these correlations, we suggest a model where the correlation between trials decays as a power law, implying that successive trials become less independent as we increase the number of trials.

In order to incorporate the correlation between trials, we define the notion of an \textit{effective number of independent trials}, denoted $k_{\text{eff}}$. This adjusts the original $k$ to account for the decay in trial independence. The correlation between trials is modeled via a power-law decay in the eigenvalues of the correlation matrix, such that the effective number of independent trials is given by
\begin{align}
\label{eq:keff}
k_{\text{eff}} = \sum_{i=1}^{k} i^{-\kappa} = H_k(\kappa)
 \underset{k\to \infty}{\approx }
\left(\frac{1}{2}-\frac{k}{\kappa-1}\right) k^{-\kappa}+\zeta (\kappa), \qquad  \kappa\in \mathbb{R}_{\geq 0}
\end{align}
where $\kappa$ is the power-law exponent governing how quickly the correlation between trials decays, $H_k(\kappa)$ is the Harmonic number, and $\zeta(\kappa)$ is the Riemann zeta function.

Thus, the probability of at least one success in $k$ trials, incorporating correlations, becomes
\begin{align}
\label{eq:pass@k}
\text{pass@k} = 
\mathcal{A}\times\left( 1 - \frac{1}{n}\sum_{i=1}^{n}p_i^{k_{\text{eff}}} \right)
=
\mathcal{A}\times
\left( 1- 
% \frac{1}{n}
% \sum_{i=1}^{n}
p^{H_k(\kappa)} \right)
\approx
\mathcal{A}\times
\left(
1-
% \frac{1}{n}\sum_{i=1}^{n}
p^{\left(\frac{1}{2}-\frac{k}{\kappa-1}\right) k^{-\kappa}+\zeta (\kappa)}
\right),
\end{align}
where $p=p_i$ is the error probability of every sample, and $k_{\text{eff}}$ accounts for correlations between trials. The result of \cref{eq:pass@k} is shown in ~\cref{fig:appx_results} as the dotted curves, which approximate the LLM behavior well for $k \gg1$. We stress that this approach should be taken as an effective description, which nevertheless manages to accurately capture the same behavior as sample correlations.

\subsection{Connection to Compute Scaling}

\begin{figure}[t!]
    \centering
    \includegraphics[width=.98\linewidth]{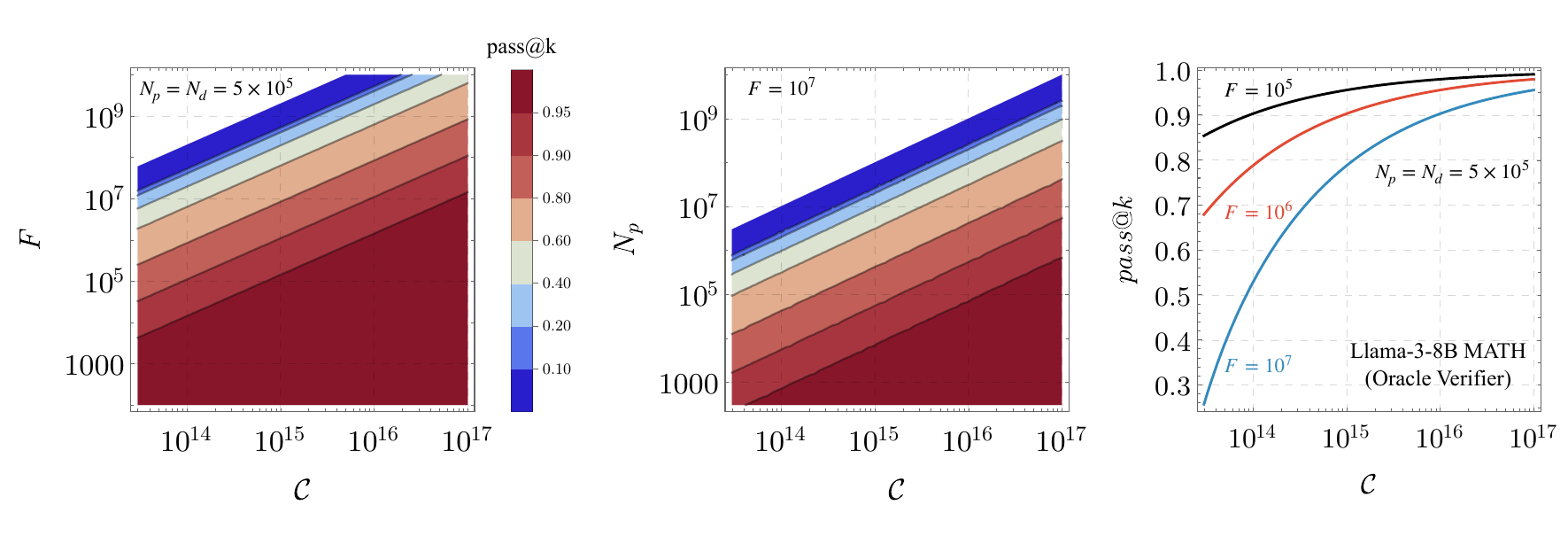} 
    \caption{{\bf{Pass@k as a function of total inference cost for Llama-3-8B MATH (Oracle Verifier)}}. 
    {\it Left and Center:} We show the pass@k metric as a function of number of total inference cost and number of FLOPS per token $F$ or number of prompt/decode tokens $N_p=N_d$ in $\log,\log$. We see that  there is a clear trade-off between total inference cost whenever keeping one of the parameters fixed, in a predictable way from~\cref{eq:cost}.
    {\it Right:} We show a slice of the contour plots for fixed $N_p=N_d$, and changing the number of FLOPS per token. 
    The parameters chosen for these figures are fitted from~\cref{eq:main_result} applied to the data taken from~\cite{brown2024largelanguagemonkeysscaling}.
}
    \label{fig:costs}
\end{figure}

Here, we would like to translate our results from the attempts variable to the inference cost. A natural proxy for the inference cost may be the number of required Floating Point Operations Per Second (FLOPS)\footnote{This may be too crude a cost metric that ignores other aspects of system efficiency \citep{DBLP:journals/corr/abs-2110-12894}, which may benefit from repeated sampling ~\citep{Juravsky2024HydragenHL,zheng2023efficiently}, as pointed out in~\cite{brown2024largelanguagemonkeysscaling}.}. 
For concreteness, we adapt the total inference cost formula suggested in~\cite{brown2024largelanguagemonkeysscaling}, given by
\begin{align}
    \text{total inference FLOPS}
    &\approx 
    \text{num prompt tokens}
    \times 
    \text{FLOPS per token}
    \\ \nonumber
    &+ 
    \text{num decoded tokens} \times 
    \text{FLOPS per token} \times 
    \text{num completions}
    \\ \nonumber
    &\to
    \mathcal{C}
    =N_{p} \times F + N_d \times F \times k,
\end{align}
where $\mathcal{C}$ is the total inference cost, $N_p,N_d$ are the number of prompt and decode tokens, respectively, and $F$ is the number of FLOPS per token.
We can convert some of our pass@k results to this metric by taking the large $k$ limit of~\cref{eq:main_result}, giving 
\begin{align}
\label{eq:cost}
    k = \left(\frac{(\mathcal{A}-\text{pass@k}) B(\alpha ,\beta )}{\mathcal{A}\Gamma (\beta )}\right)^{-1/\beta }
    \to
    \text{Coverage}(\mathcal{C})
    \approx
    \mathcal{A}\times\left(
    1-\frac{\Gamma (\beta ) }{B(\alpha ,\beta )}
    {\left(\frac{\bar{\mathcal{C}}-N_p }{N_d }\right)^{-\beta }}
    \right),
\end{align}
where $\bar{\mathcal{C}}\equiv \mathcal{C}/F$ is the normalized total inference cost.
Keeping all uncontrollable parameters fixed, such as the number of FLOPS per token fixed, we see from~\cref{fig:costs} that it is worthwhile to reduce the number of prompting and decoding tokens or increase the number of attempts, up to the minimal amount of total inference compute required to reach a target coverage. The numbers in~\cref{fig:costs} are chosen to mimic the results found in~\cite{brown2024largelanguagemonkeysscaling}, and are only meant to give an illustration of the functional behavior of \cref{eq:cost}.
\\
Alternatively, we can phrase \cref{eq:cost} in terms of the inference loss as
\begin{align}
    \mathcal{L}_{\text{inference}}(\mathcal{C})
    \approx
    \mathcal{A}\times
    \frac{\Gamma (\beta ) }{B(\alpha ,\beta )}
    {\left(\frac{\bar{\mathcal{C}}-N_p }{N_d }\right)^{-\beta }} , 
\end{align}
which demonstrates the power law decay of the inference loss with total inference cost, depending on the value of $\beta$.

\section{Experiments on a Simple Generative Model}
\label{sec:experiments}

\begin{figure}[t!]
    \centering
    \includegraphics[width=1\linewidth]{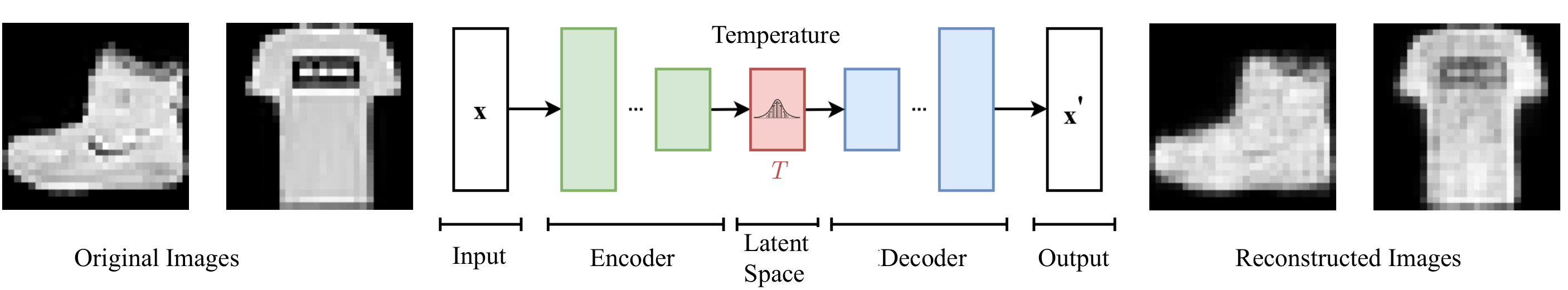} 
    \caption{{ \bf{Visualization of the task described in~\cref{sec:experiments}}}. Here, a VAE is tasked with generating samples from its training data, where a "failure" occurs when the reconstruction error falls above a certain threshold $\epsilon$.}
    \label{fig:VAE_exp}
\end{figure}

To further validate some of our analytical understanding in a controllable setting, we perform a series of experiments in which we train a simple generative model to reconstruct images taken from Fashion-MNIST~\citep{xiao2017fashionmnistnovelimagedataset}.  
Our goal is to connect the theoretical memorizing model and the behavior of more complex generative models by attempting to accurately reconstruct "memorized" examples, heuristically shown in~\cref{fig:VAE_exp}.

To do this, we train a variational autoencoder (VAE) with a temperature parameter to study how errors propagate over multiple trials and to compare empirical pass@k with theoretical predictions under correlated trials. We refer to this as the {\it VAE reconstruction task}.

To quantify the error probability of the model over multiple samples, we define the error per sample using the norm of the difference between the reconstructed and original image:
\begin{align}
\text{error}(i) = \frac{\|\hat{y}_i - y_i\|}{\|y_i\|}.
\end{align}
Here, $y_i$ represents the original image, and $\hat{y}_i$ is the reconstruction. This per-sample error metric allows us to define success or failure at the sample level, where a trial is considered successful if the reconstruction error falls below a threshold $\epsilon$.

% \vspace{-10pt}
To empirically calculate pass@k, we sample multiple reconstructions from the VAE for each input sample. For each sample, we conduct $k$ trials and calculate whether at least one trial resulted in a reconstruction with error less than some chosen threshold value $\epsilon \in [0,1]$. Pass@k is then computed as the fraction of samples for which the model succeeded to reconstruct at least once in $k$ trials.

\begin{figure}[t!]
    \centering
        \includegraphics[width=.95\linewidth]{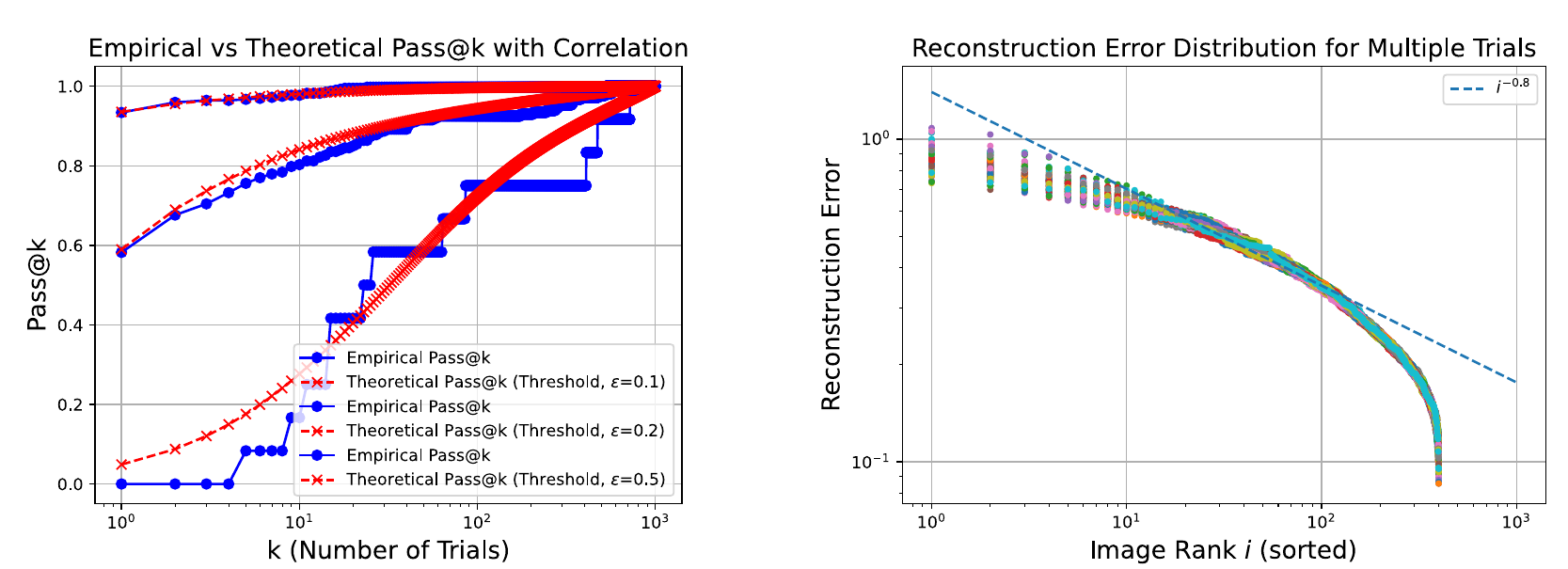} 
    \caption{ {\bf Results for the VAE reconstruction task, compared with semi-analytical predictions}.
    {\it Left:} The pass@k metric as a function of number of attempts $k$, for different threshold values, with temperature $T=1.1$. The curves have been normalized to asymptote at 1 for visual clarity.
    {\it Right:} The reconstruction error behavior across multiple trials, indicated by different colors. The errors obey a quasi power law behavior.   }
    \vspace{-0.4cm}
    \label{fig:VAE_results}
\end{figure}
\begin{figure}[t!]
    \centering
    \includegraphics[width=0.42\linewidth]{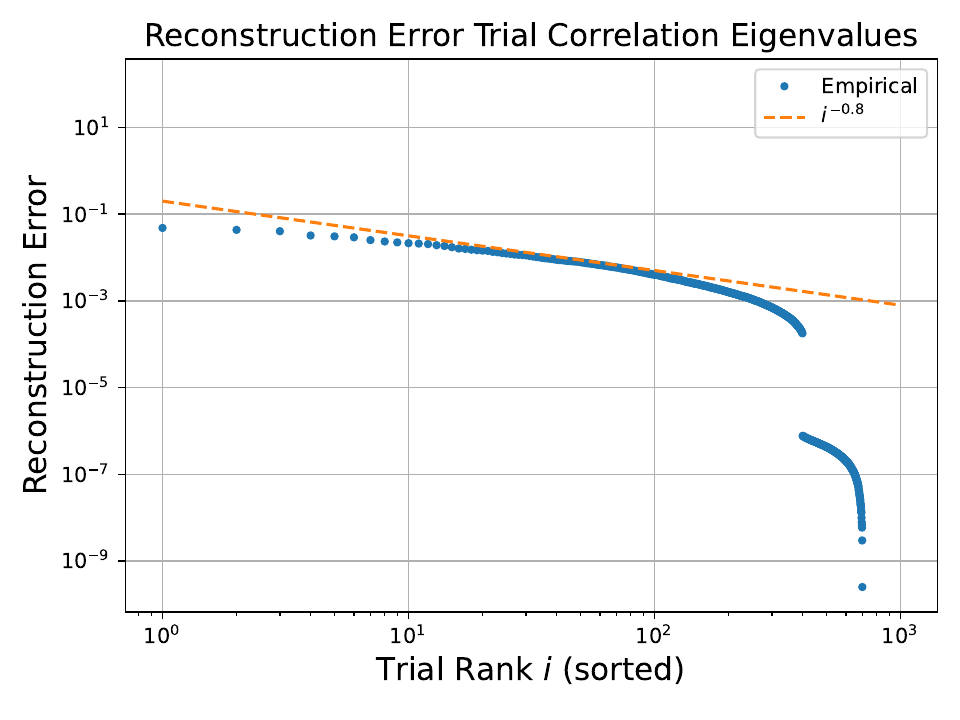}
    \includegraphics[width=0.45\linewidth]{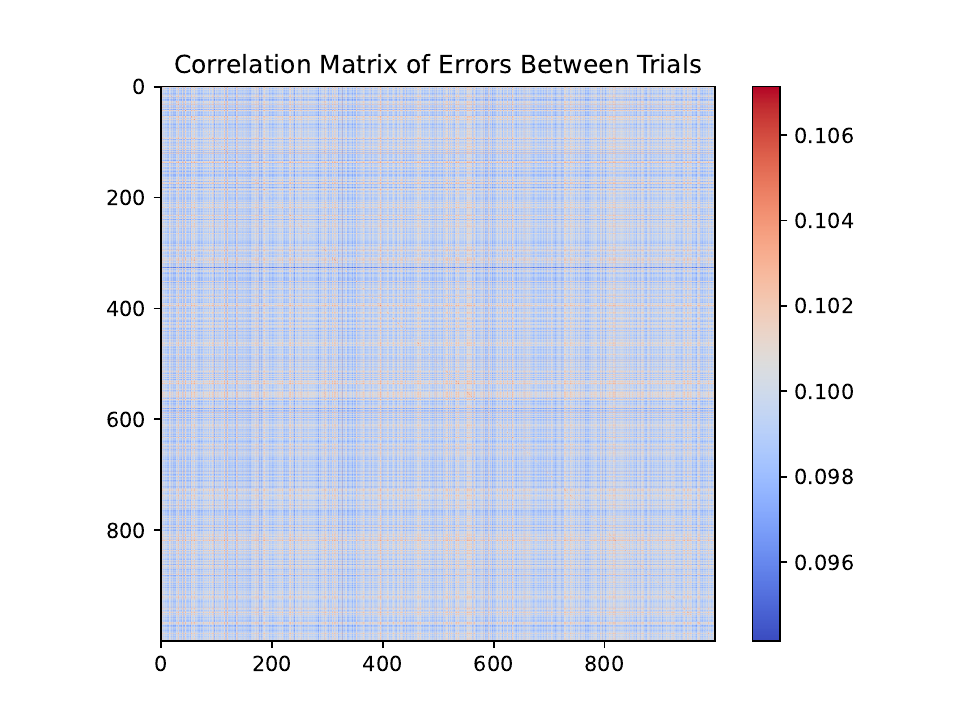}   
    \caption{{\bf Correlation matrix for errors across different trials for the VAE reconstruction task.} {\it Left:} The eigenvalues of the correlation matrix follow a power law decay with the number of trials. {\it Right:} Visualization of the correlation matrix itself shows clear correlations between different trials for $k=1000$.}
    \label{fig:trial_correlations}
\end{figure}

In~\cref{fig:VAE_results}, we show the pass@k results for the VAE reconstruction task, for different threshold $\epsilon$ values (left), as well as the reconstruction error distribution for multiple attempts (right). The theoretical predictions for the pass@k curves are shown for the effective $k$ approach, given in~\cref{eq:pass@k}, and approximate the VAE results well. Here, $p_i$ is computed empirically by taking the distribution at the maximal $k$ and $\kappa$ is taken from the correlation matrix in~\cref{fig:trial_correlations}.

To complete the picture, we empirically confirm that the assumption of independent trials is indeed violated, as trials are effectively correlated. To capture this effect, we compute the correlation matrix for the errors across trials $\epsilon_{k k'}$ as
\begin{align}
    \epsilon_{k k'}
    =
    \frac{1}{n}\sum_i \text{error}_{i,k} 
    \times
    \text{error}_{i,k'}.
\end{align}
The eigenvalues of the correlation matrix decay as a power law, suggesting that the effective number of independent trials diminishes as $k$ increases, which is clearly depicted in~\cref{fig:trial_correlations} (left).

\section{Conclusion}

In this paper, we have proposed a simple statistical explanation for a so-called inference scaling law, which describes the scaling of the coverage (pass@$k$) metric with number of repeated attempts for generative models. 
We presented two possible models which lead to inference scaling: One based on introducing a sample space distribution of "easy" and "difficult" problems, and the other on an effective Zipf-like correlation structure between trials.
Using these simple constructions, we were able to derive analytical predictions for the pass@k, as well the test loss as a function of repeated attempts, which we dubbed the inference loss. 
We then verified our predictions empirically both through previous experimental results for LLMs and for a simple generative VAE construction.

We stress that the merit of our construction is in its simplicity, and there are many other models who can give rise to the same functional behavior. We view this as a positive rather than a negative, since it means that this simple model captures a universal behavior, which should not depend much on the modeling itself. For instance, another way to arrive at a similar scaling law would be to choose a different modeling for the failure distribution, based perhaps on program length, and introducing the notion of a distribution of program lengths corresponding to different samples, similar to~\cite{ringel2018criticalpercolationframeworkanalyze}. In the end, this type of construction will have a similar interpretation in terms of task complexity w.r.t the model.

We believe our toy model offers a simple yet effective phenomenological framework for understanding how inference quality improves with more opportunities to predict correctly. Future work could extend this framework to more complex models, including applying similar methodology as \cite{maloney2022solvable} to generalized linear regression, kernel regression and neural networks, and investigate how it interacts with existing scaling laws based on model size and training data.

\section{Acknowledgements}

We thank Yohai Bar-Sinai, Nadav Outmezguine, Zohar Ringel and Antonio Sclocchi for fruitful discussions. NL is supported by the EPFL AI4science program.

\bibliographystyle{unsrtnat}
\bibliography{bib}
% \begin{thebibliography}{9}

% \bibitem{kaplan2020scaling}
% Kaplan, J., McCandlish, S., Henighan, T., et al. (2020). \textit{Scaling Laws for Neural Language Models}. arXiv preprint arXiv:2001.08361.

% \bibitem{henighan2020scaling}
% Henighan, T., Kaplan, J., et al. (2020). \textit{Scaling Laws for Autoregressive Generative Modeling}. arXiv preprint arXiv:2010.14701.

% \bibitem{hutter2001theory}
% Hutter, M. (2001). \textit{A Theory of Universal Artificial Intelligence based on Algorithmic Complexity}. arXiv preprint arXiv:cs/0004001.

% \end{thebibliography}
\newpage
\appendix

\section{Experimental Setup}

\begin{figure}[ht!]
    \centering
    \includegraphics[width=1.\linewidth]{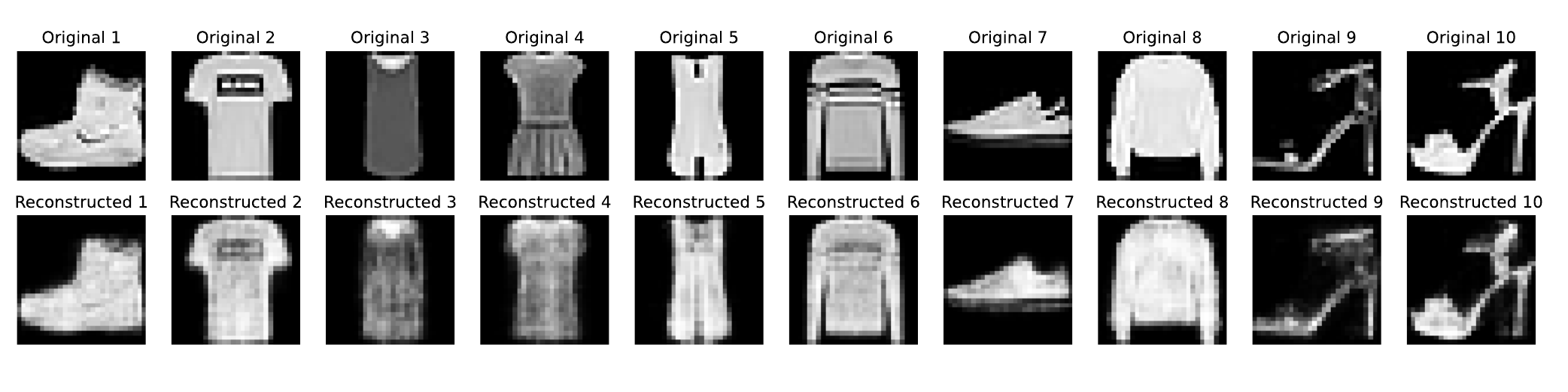}
    \caption{{\bf Examples of original and reconstructed samples from the VAE reconstruction task.} }
    \label{fig:images}
\end{figure}

Here, we specify the experimental procedure used for the analysis in~\cref{sec:experiments}.

% \subsection{Model Architecture}
We utilize a Variational Autoencoder (VAE) with the following architectural details:
\setlength{\itemsep}{0pt} % Adjusts space between items
\setlength{\parskip}{0pt} % Adjusts space between paragraphs
\setlength{\topsep}{0pt}  % Adjusts space between the list and surrounding content
\begin{itemize}
    \item \textbf{Input dimension:} \( 28 \times 28 = 784 \), corresponding to the flattened pixel values of the Fashion MNIST dataset.
    \item \textbf{Hidden dimension:} 400.
    \item \textbf{Latent dimension:} 20, controlling the bottleneck for information in the latent space.
    \item \textbf{Decoder:} The decoder reconstructs the original input through two fully connected layers, outputting a 784-dimensional vector followed by a sigmoid activation to ensure pixel values remain between 0 and 1.
    \item \textbf{Temperature parameter:} A temperature parameter \( T = 1.1 \) is applied during the reparameterization step to control the variance of the latent variables, allowing us to model uncertainty in the latent space more effectively.
\end{itemize}

% \subsection{Training Procedure}
The VAE was trained on the first 400 samples from the Fashion MNIST dataset. The loss function combines binary cross-entropy for reconstruction and the Kullback-Leibler divergence to regularize the latent variables. We ran the training for 1000 epochs using the Adam optimizer with a learning rate of \( 1 \times 10^{-3} \).

To provide qualitative insights into the model's performance, we visualize several input samples from the Fashion-MNIST dataset along with their corresponding reconstructions in~\cref{fig:images}. This allows us to inspect both successful and failed reconstructions, and examine the types of errors the model makes.

% \subsection{Different trials are effectively correlated}

% The results for the error correlation matrix between trials for the VAE reconstruction task, shown in~\cref{fig:trial_correlations}, was obtained by considering the error matrix $\text{error}_{i,k}$ where $i$ is the sample index and $k$ is the trial index.
% The correlation matrix in trial space $\epsilon_{k k'}$ is defined as
% \begin{align}
%     \epsilon_{k k'}
%     =
%     \frac{1}{n}\sum_i \text{error}_{i,k} 
%     \times
%     \text{error}_{i,k'}.
% \end{align}
% In~\cref{fig:trial_correlations} we show the correlation matrix (right) and its eigenvalues (left).
% Clearly, the eigenvalues of the error correlation matrix for $k=1000$ trials follow a power law decay with the number of trials, as described in the main text.

% \begin{figure}[t]
%     \centering
%     \includegraphics[width=0.40\linewidth]{Figs/correlations_EVS.pdf}
%     \includegraphics[width=0.42\linewidth]{Figs/correlations.pdf}    
%     \caption{{\bf Correlation matrix for errors across different trials for the VAE reconstruction task.} {\it Left:} The eigenvalues of the correlation matrix follow a power law decay with the number of trials. {\it Right:} Visualization of the correlation matrix itself shows clear correlations between different trials.}
%     \label{fig:trial_correlations}
% \end{figure}

\end{document}